\documentclass[11pt]{article}
\usepackage{coling2020}
\usepackage{times}
\usepackage{url}
\usepackage{latexsym}
\usepackage{amssymb}
\usepackage{color}
\usepackage{mathtools}
\usepackage{proof, lscape}
\usepackage{gb4e}
\usepackage{multirow}

\newcommand{\bhline}[1]{\noalign{\hrule height #1}}

\newcommand{\PredicateFont}[1]{\mathsf{#1}}
\newcommand{\LF}[1]{\ensuremath{\PredicateFont{#1}}}
\newcommand{\fun}[2]{\ensuremath{\PredicateFont{#1}({#2})}}
\newcommand{\funt}[3]{\ensuremath{\PredicateFont{#1}({#2},{#3})}}
\newcommand{\subj}[2]{\ensuremath{\PredicateFont{subj}(#1) = {#2}}}
\newcommand{\acc}[2]{\ensuremath{\PredicateFont{obj}(#1) = {#2}}}
\newcommand{\funtheta}[2]{\ensuremath{\theta_{\PredicateFont{#1}}(\PredicateFont{#2})}}
\newcommand{\posth}[1]{\ensuremath{\theta_{\PredicateFont{#1}}}}

\newcommand{\LabelYes}{\textit{yes}}
\newcommand{\LabelUnk}{\textit{unknown}}
\newcommand{\LabelNo}{\textit{no}}

\colingfinalcopy 

\title{Combining Event Semantics and Degree Semantics \\for Natural Language Inference}

\author{Izumi Haruta$^{1}$ \\
        \texttt{haruta.izumi@is.ocha.ac.jp}
        \AND
        Koji Mineshima$^{2}$\\ 
        \texttt{minesima@abelard.flet.keio.ac.jp}
        \And
        Daisuke Bekki$^{1}$ \\ 
        \texttt{bekki@is.ocha.ac.jp}
        \AND
        $^{1}$\mbox{\rm Ochanomizu University, Tokyo, Japan} \\
        $^{2}$\mbox{\rm Keio University, Tokyo, Japan}
        }

\date{}

\begin{document}
\maketitle
\begin{abstract}

In formal semantics, there are two well-developed semantic frameworks:
event semantics, which treats verbs and adverbial modifiers using the notion of \textit{event},
and degree semantics, which analyzes adjectives and comparatives using the notion of \textit{degree}.
However, it is not obvious whether these frameworks can be combined to handle cases in which the phenomena in question are interacting with each other. 
Here, we study this issue by focusing on natural language inference (NLI).
We implement a logic-based NLI system that combines event semantics and degree semantics
and their interaction with lexical knowledge.
We evaluate the system on various NLI datasets containing linguistically challenging problems.
The results show that the system achieves high accuracies on these datasets
in comparison with previous logic-based systems and deep-learning-based systems.
This suggests that the two semantic frameworks can be combined consistently
to handle various combinations of linguistic phenomena without compromising the advantage of either framework.
\end{abstract}

\section{Introduction}
\label{sec:intro}

Since \newcite{Montague1970-MONUG}, formal compositional semantics has provided successful accounts of
linguistic phenomena using logical expressions along with syntactic structures.
In recent years, with the development of wide-coverage parsers such as Combinatory Categorial Grammar (CCG) parsers~\cite{Clark2007},
some of the formal theories have been implemented 
as robust computational semantics~\cite{bos2008wide,mineshima2015higher,abzianidze2016natural}
and have been applied to natural language inference (NLI), 
which is the task of determining whether a text entails a hypothesis.
This paper attempts to push forward this paradigm by focusing on 
the interaction of two semantic frameworks, namely, event semantics and degree semantics.

Generally speaking, research in linguistic formal semantics has tended to
focus on creating an in-depth theory for a variety of linguistic phenomena,
such as quantifiers, adjectives, comparatives, and tense.
However, it is often not obvious whether these independent theories can be combined
and extended to cases in which the phenomena in question interact.
Thus, event semantics has been developed largely to account for the semantics of verb phrases and adverbial modifiers~\cite{Davidson67,parsons1990events}.
In contrast, degree semantics 
provides an analysis of gradable expressions such as adjectives and comparatives, using the notion of degree~\cite{cresswell1976semantics,stechow1984comparing,kennedy97}.
Although each theory has been elaborated for its own sake,
it is not clear how to combine these two and handle expressions that require
the application of both theories, such as the comparative form of adverbs,
though it is a necessary step for analyzing real texts.

The computational modeling of compositional semantic theories
mentioned above enables us to precisely compute the predictions of each theory.
In addition, their application to NLI tasks provides a systematic way of
evaluating a formal semantic theory.
Building on the previous logic-based approaches to NLI,
we present a logic-based NLI system that combines event semantics and degree semantics,
and evaluate the system on various datasets to check whether the system can perform linguistically challenging inferences
without compromising the accuracy of each basic theory.

More specifically, 
we build on the system presented in \newcite{haruta-etal-2020-logical},
which implements degree semantics for comparatives and generalized quantifiers.
The system is limited, however, in that it does not implement event semantics, and hence
does not handle inferences with adverbs and related constructions.
Also, it only covers a small portion of the generalized quantifiers discussed in the linguistics literature
and does not handle inferences requiring lexical knowledge, thus being confined to purely logical inferences.
To test the compatibility of event semantics and degree semantics,
we add a layer of event semantics and lexical knowledge to that system and attempt to 
broaden its empirical coverage consistently.

To evaluate the system, we assess its capacity to handle
(i) a set of logical compositional inferences and (ii) their interaction with lexical knowledge.
For (i), we use FraCaS~\cite{cooper1994fracas}, which contains various semantically complex inferences
and CAD~\cite{haruta-etal-2020-logical}, which contains complex inferences with adjectives and comparatives.
To our knowledge, there is no linguistically controlled dataset that contains inferences with adverbs,
so we create a set of logical inferences with adverbs and related constructions
and use it to evaluate the system.
For (ii), we use MED~\cite{yanaka-etal-2019-neural}, which contains
problems with monotonicity inferences and lexical knowledge, and SICK~\cite{marelli-etal-2014-sick}, which focuses on lexical inferences combined with linguistic phenomena such as negation and quantifiers.
In addition, we use HANS~\cite{mccoy-etal-2019-right}, 
which is designed to probe the capacity of NLI models based on deep learning (DL)
and contains structural inferences, including those concerning adjectives and adverbs.
The results show that our system achieves high accuracies across various datasets,
supporting our claim that event semantics and degree semantics can be effectively combined for NLI.

\section{System architecture}
\label{sec:system}

For the implementation of an NLI system\footnote{GitHub repository with code and data: \url{https://github.com/izumi-h/ccgcomp}}, we follow the basic architecture of \newcite{haruta-etal-2020-logical}.
Input sentences (i.e., a set of premises $P_1, \ldots, P_n$ and a hypothesis $H$)
are mapped to CCG derivation trees using off-the-shelf CCG parsers.
To accommodate output derivation trees
in formal semantic analysis,
various tree transformations are applied.
Then, using a set of semantic templates that assign lambda-expressions to CCG categories~\cite{Steedman2000},
the output trees are mapped to logical forms,
which are formulas of first-order logic (FOL) with equality and arithmetic operations.
This provides a set of formulas $P_1', \ldots, P_n'$ for the premises and $H'$ for the hypothesis.
Then, an FOL theorem prover tries to prove  $P_1' \wedge \cdots \wedge P_n' \to H'$.
If it is successful, the system outputs \textit{yes} (entailment).
Otherwise, the system tries to prove $P_1' \wedge \cdots \wedge P_n' \to \neg H'$
and outputs \textit{no} (contradiction).
If both attempts fail, the system outputs \textit{unknown} (neutral).
For our purposes, we add two components to this basic architecture:
(i) we extend the set of the semantic template to accommodate event semantics
and various types of generalized quantifiers,
and (ii) we add a mechanism to insert lexical knowledge before theorem proving.

\begin{table*}[t]
  \centering
  \scalebox{0.7}{$
  {\renewcommand\arraystretch{1.1}
  \begin{tabular}{l|l}\bhline{1.3pt}
    \textbf{Example} &
    \textbf{Logical form}\\
    \hline\hline
    John shouted \textit{loudly}. & $\exists e(\fun{shout}{e} \wedge (\subj{e}{\LF{john}}) \wedge \funt{loud}{e}{\posth{loud}})$\\
    Ann studied English \textit{very hard}. & $\exists e(\fun{study}{e} \wedge (\subj{e}{\LF{ann}}) \wedge (\acc{e}{\LF{english}}) \wedge \exists \delta(\funt{hard}{e}{\delta} \wedge (\posth{hard} < \delta)))$\\
    Jim sings \textit{better than} Mary. & 
    $\exists e_1 \exists e_2(\fun{sing}{e_1} \wedge (\subj{e_1}{\LF{jim}}) \wedge \fun{sing}{e_2} \wedge (\subj{e_2}{\LF{mary}}) \wedge \exists \delta(\funt{good}{e_1}{\delta} \wedge \neg\funt{good}{e_2}{\delta}))$\\
    Bob drives \textit{as carefully} as John. & 
    $\exists e_1 \exists e_2(\fun{drive}{e_1} \wedge (\subj{e_1}{\LF{bob}}) \wedge \fun{drive}{e_2} \wedge (\subj{e_2}{\LF{john}}) \wedge \forall \delta(\funt{careful}{e_2}{\delta} \to \funt{careful}{e_1}{\delta}))$\\
    \bhline{1.3pt}
    \end{tabular}
    }
    $}
    \caption{Logical forms of adverbs and their comparative and equative forms}
    \label{tab:adverb}
\end{table*}

\paragraph{Combining event semantics and degree semantics}

We use standard neo-Davidsonian event semantics~\cite{parsons1990events},
which analyzes sentences as involving quantification over events.
For instance, the sentence \textit{John ran} is analyzed as
$\exists e(\fun{run}{e} \wedge (\subj{e}{\LF{john}}))$, where $\mathsf{subj}$ is a function term
that associates an event to its participant (subject).
A sentence containing an adverb, for example, \textit{John ran slowly},
is analyzed as
$\exists e (\fun{run}{e} \wedge (\subj{e}{\LF{john}}) \wedge \fun{slowly}{e})$,
where the adverb \textit{slowly} acts as a predicate of an event.
This allows us to derive an inference from \textit{John ran slowly} to \textit{John ran},
that is, an inference to drop adverbial phrases.

For degree semantics, we use the one presented in \newcite{haruta-etal-2020-logical}.
It analyzes a gradable adjective \textit{tall} as a binary predicate $\LF{tall}(x,\delta)$ holding of
an entity $x$ and a degree $\delta$~\cite{cresswell1976semantics}.
Thus, the sentence \textit{Chris is 5 feet tall} is mapped to the logical form $\funt{tall}{\LF{chris}}{\LF{5~feet}}$.
When a degree expression like \textit{5 feet} is absent, 
$\delta$ is set to a default value (for which we use a fixed constant);
e.g., \textit{Chris is tall} is mapped to $\funt{tall}{\LF{chris}}{\posth{tall}}$.
Comparative expressions are analyzed in terms of first-order logic, using the so-called A-not-A analysis~\cite{seuren1973comparative,klein1982interpretation,schwarzschild2008semantics}; for example,
\textit{Chris is taller than Alex} is analyzed as $\exists \delta(\funt{tall}{\LF{chris}}{\delta} \wedge \neg\funt{tall}{\LF{alex}}{\delta})$, which asserts that there is a degree $\delta$ of tallness
that Chris satisfies but Alex does not. See \newcite{haruta-etal-2020-logical} for more detail.
This analysis can be naturally extended to adverbial phrases and their comparative forms.
Table \ref{tab:adverb} shows logical forms of basic constructions, where adverbs like \textit{slowly} are treated as binary predicates
of an event and a degree.

\begin{table*}[t]
  \centering
  \scalebox{0.67}{$
  {\renewcommand\arraystretch{1.1}
  \begin{tabular}{l|l|l}
  \bhline{1.3pt}
    \textbf{Type} & \textbf{Example} &
    \textbf{Logical form}\\
    \hline\hline
    upward & \textit{Many} people cried. & $\exists x (\fun{people}{x} \wedge \funt{many}{x}{\funtheta{many}{person}} \wedge \exists e(\fun{cry}{e} \wedge (\subj{e}{x})))$\\
    downward & \textit{Less} than five students laughed. & $\neg \exists x(\fun{student}{x} \wedge \funt{many}{x}{5} \wedge \exists e(\fun{laugh}{e} \wedge (\subj{e}{x})))$\\
    non-monotone & \textit{Exactly} eleven boys play soccer. & $\exists x (\fun{boy}{x} \wedge \funt{many}{x}{11} \wedge \exists e(\fun{play}{e} \wedge (\subj{e}{x}) \wedge (\acc{e}{\LF{soccer}})))$ \\
    & & $\wedge \forall x \forall \delta (\fun{boy}{x} \wedge \funt{many}{x}{\delta} \wedge \exists e(\fun{play}{e} \wedge (\subj{e}{x}) \wedge (\acc{e}{\LF{soccer}})) \to (\delta < 12))$\\
    \bhline{1.3pt}
    \end{tabular}
    }
    $}
    \caption{Logical forms of upward, downward, and non-monotonic generalized quantifiers}
    \label{tab:mono}
\end{table*}

\paragraph{Monotonicity and generalized quantifiers}
In degree semantics, generalized quantifiers such as \textit{many}
can be analyzed not as higher-order expressions (as in Montague's tradition)
but as two-place predicates such as $\LF{many}(x,n)$,
which reads ``a composite entity $x$ consists of at least $n$ individuals''~\cite{hackl2000comparative,rett2018semantics}.
\newcite{haruta-etal-2020-logical} implements this semantics in their system but does not
deal with downward quantifiers (e.g., \textit{few} and \textit{less than five})
or non-monotonic quantifiers (e.g., \textit{exactly five}).
It is known that these two types of quantifiers pose a problem for the compositional treatment of quantifiers as predicates~\cite{van1986essays,buccola2016}.
The problem is that downward and non-monotonic quantifiers need to take scope over the entire clause.
For example, in the case of the downward quantifier \textit{less than} in \textit{Less than five students laughed},
the negation needs to take scope over the entire clause, as shown in Table \ref{tab:mono}.
To solve this problem, we add syntactic features to CCG categories in the derivation trees
in the post-processing process.
Downward quantifiers such as \textit{few} and \textit{less than five}
are assigned the category $N/N$ (which is the same as an adjective like \textit{tall}) in CCG parsers.
We modify this to $N_{down}/N$, which triggers the derivation in which the negation takes the outermost scope.
Similarly, we assign syntactic categories like $N_{nm}/N$ to non-monotonic quantifiers such as \textit{exactly} and \textit{only}.
This enables us to derive the desired logical forms as in Table \ref{tab:mono}.

\paragraph{Insertion of lexical knowledge}

To test the compatibility of logical inferences and inferences involving lexical knowledge,
we implement a mechanism to search for useful axioms drawn from knowledge bases
before the process of theorem proving.
The strategy is similar to the one used in previous studies~\cite{martinez-gomez-etal-2017-demand} in which
the system searches for lexical relations 
from WordNet~\cite{miller1995wordnet} and VerbOcean~\cite{chklovski-pantel-2004-verbocean}.
More specifically, for each predicate $F$ appearing in the set of formulas for given premises,
if there is a predicate $G$ appearing in the formula for the hypothesis
such that (i) $F$ and $G$ have the same semantic type (e.g., the type of predicate of events) and 
(ii) $F$ has a lexical relationship with $G$,
then we add an axiom
of the relevant form
depending on the type of lexical relation.
Following \newcite{martinez-gomez-etal-2017-demand},
we use a total of seven relationships
and add the corresponding axioms, as shown in Table \ref{tab:lexax}.

\begin{table}[t]
    \centering
    \scalebox{0.75}{$\displaystyle
    {\renewcommand\arraystretch{1.1}
    \begin{tabular}{r|c|c|c|c}
        \textbf{Lexical relationship} &
         antonym & hypernym & hyponym & synonym, similar, inflection, derivation\\
         \hline
         \textbf{Axiom} &
         $\forall x(F(x) \to \neg G(x))$ & $\forall x(F(x) \to G(x))$ & $\forall x(G(x) \to F(x))$ & $\forall x(F(x) \leftrightarrow G(x))$\\
    \end{tabular}
    }
    $}
    \caption{Correspondence between the lexical relationships and the forms of inserted axioms}
    \label{tab:lexax}
\end{table}

\section{Experiments}
\label{sec:experiment}

\begin{table}[t]
    \centering
    \scalebox{0.63}{$\displaystyle
  	\begin{tabular}{c|c|c||l|c}\bhline{1.3pt}
        Dataset & Label & ID & Example (premises and hypothesis) & Gold label\\
    	\hline\hline
    	\multirow{3}{*}{FraCaS} & \multirow{3}{*}{\textsl{Com}} & & $P_1$: ITEL won more orders than APCOM lost.&\\
    	& & 241 & $P_2$: APCOM lost ten orders. & Yes\\
    	& & & $H$: ITEL won at least eleven orders. &\\
    	\hline
    	\multirow{4}{*}{MED} & \multirow{2}{*}{\textsl{gq}} & 
        \multirow{2}{*}{485} & $P_1$: Exactly 12 aliens threw some tennis balls. & \multirow{2}{*}{Unknown}\\
    	& & & $H$: Exactly 12 aliens threw some balls. & \\
    	\cline{2-5}
    	 & \multirow{2}{*}{\textsl{gqlex}} & \multirow{2}{*}{176} & $P_1$: Few aliens saw birds. & \multirow{2}{*}{Yes}\\
    	& & & $H$: Few aliens saw doves. & \\
    	\hline
    	\multirow{2}{*}{SICK} & \multirow{2}{*}{--} & \multirow{2}{*}{1357} & $P_1$: A puppy is repeatedly rolling from side to side on its back. & \multirow{2}{*}{Yes}\\
    	& & & $H$: A dog is rolling from side to side. & \\
    	\hline
    	\multirow{2}{*}{HANS} & \multirow{2}{*}{\textsl{constituent}} & \multirow{2}{*}{23991} & $P_1$: The actors contacted the president, or the lawyers recommended the managers. & \multirow{2}{*}{Unknown}\\
    	& & & $H$: The lawyers recommended the managers. & \\
    	\hline
        \multirow{5}{*}{CAD+} & \multirow{5}{*}{\textsl{MA}} & \multirow{2}{*}{115} & $P_1$: Exactly seven students smiled. & \multirow{2}{*}{Yes}\\
    	& & & $H$: At most nine students smiled. &\\
    	\cline{3-5}
        & & \multirow{3}{*}{157} & $P_1$: Ann runs as fast as Luis does. & \\
    	& & & $P_2$: Ann runs slowly. & No\\
    	& & & $H$: Luis runs fast. &\\
    	\bhline{1.3pt}
  	\end{tabular}
  	$}
    \caption{Examples of entailment problems from the FraCaS, MED, SICK, HANS, and CAD+ datasets}
    \label{tab:data}
\end{table}

\paragraph{Experimental settings}
We use three CCG parsers, namely,
C\&C~\cite{Clark2007}, EasyCCG~\cite{lewis2014ccg}, and depccg~\cite{yoshikawa-etal-2017-ccg},
for CCG parsing,
and we use Tsurgeon \cite{levy-andrew-2006-tregex} for tree transformation.
For CCG parsing and tree transformation, we use the same setting as in \newcite{haruta-etal-2020-logical}.
We use a set of semantic templates for mapping CCG trees to logical forms.
The templates are specified for 528 categories and 138 lemmas in total.
The Tsurgeon script has 126 clauses for tree mapping.
For POS tagging, we use
the C\&C POS tagger for C\&C
and spaCy\footnote{
\url{https://github.com/explosion/spaCy}
} for EasyCCG and depccg.
We use the FOL prover Vampire\footnote{\url{https://github.com/vprover/vampire}} for theorem proving.
    
\paragraph{Datasets} 
For evaluation, we use five datasets.
Table \ref{tab:data} shows some examples.
(1) FraCaS has nine sections,
of which we use four:
Generalized Quantifiers (\textsl{GQ}),
Adjectives (\textsl{Adj}), Comparatives (\textsl{Com}), and Attitudes (\textsl{Att}).
(2) MED
collects inferences with generalized quantifiers.
We use a portion of the dataset taken from linguistics papers, which are divided into
non-lexical inferences (\textsl{gq}: 498 problems) and inferences involving lexical knowledge (\textsl{gqlex}: 691 problems).
(3) For SICK, we use the 
2014 version of SemEval~\cite{marelli-etal-2014-sick}.
(4) For HANS,
we randomly choose 1002 problems labelled as \textit{constituent}, \textit{lexical overlap}, and \textit{subsequence} from the entire test set (30,000 problems), which
are divided into entailment (\LabelYes) and non-entailment (\LabelUnk) problems.
(5) CAD+ has two sections.
CAD~\cite{haruta-etal-2020-logical} contains 105 inference problems concerning adjectives and comparatives, which are linguistically challenging but missing from FraCaS.
We also create a new set of problems, called \textsl{MA} for monotonicity and adverbial phrases,
which follows the patterns in CAD.
It has 134 problems in total.
Of these 134 problems, 69 are single-premise problems, and 65 are multi-premise problems.
The distribution of gold answer labels is (\LabelYes/\LabelNo/\LabelUnk) = (57/36/41).

\begin{table}[t]
    \centering
    \scalebox{0.7}{$\displaystyle
	    \begin{tabular}{l|cccc}
	        \bhline{1.3pt}
	        \multicolumn{5}{l}{FraCaS}\\
	        \hline\hline
	        Section & \textsl{GQ} & \textsl{Adj} & \textsl{Com} & \textsl{Att}\\
	        \hline
	        \#All & 73 & 22 & 31 & 13\\
	        \hline
	        Maj & .49 & .41 & .61 & .62\\
	        \hline
	        \texttt{RB} & .73 & .45 & .52 & .69\\
	        \hline
	        \texttt{MN} & .77 & .68 & .48 & .77\\
	        \texttt{LP} &.93 & .73 & -- & \textbf{.92}\\
	        \texttt{HR} & .95 & \textbf{.95} & .84 & --\\
	        Ours & .97 & .82 & \textbf{.90} & \textbf{.92}\\
	        +rule & \textbf{.99} & \textbf{.95} & \textbf{.90} & \textbf{.92}\\
            \bhline{1.3pt}
  	    \end{tabular}
    \hspace{2em}
        \begin{tabular}{l|cc}
	        \bhline{1.3pt}
	        \multicolumn{3}{l}{MED}\\
	        \hline\hline
	        Label & \textsl{gq} & \textsl{gqlex}\\
	        \hline
	        \#All & 498 & 691\\
	        \hline
	        Maj & .58 & .63\\
	        \hline
	        \texttt{BERT} & .56 & .58\\
	        \texttt{BERT+} & .54 & .68\\
	        \texttt{RB} & .57 & .55 \\
	        \hline
	        \texttt{HR} & .84 & --\\
	        Ours & \textbf{.96} & \textbf{.92}\\
            \bhline{1.3pt}
        \end{tabular}
     \hspace{2em}
        \begin{tabular}{l|c}
	        \bhline{1.3pt}
	        \multicolumn{2}{l}{SICK}\\
	        \hline\hline
	        \#All & 4927\\
	        \hline
	        Maj & .57\\
	        \hline
	        \texttt{RB} & .56\\
	        \hline
	        \texttt{LP} & .81 \\
	        \texttt{MG} & \textbf{.83} \\
	        Ours & .82\\
            \bhline{1.3pt}
        \end{tabular}
    \hspace{2em}
    \centering
        \begin{tabular}{l|c|c}
	        \bhline{1.3pt}
	        \multicolumn{3}{l}{HANS}\\
	        \hline\hline
	        Gold & \LabelYes & \LabelUnk\\
	        \hline 
	        \#All & 501 & 501\\
	        \hline
	        Maj & .50 & .50\\
	        \hline
	        \texttt{RB} & \textbf{1.0} & .56\\
	        \hline
	        Ours & .97 & \textbf{.78}\\
            \bhline{1.3pt}
        \end{tabular}
    \hspace{2em}
    \centering
        \begin{tabular}{l|cc}
	        \bhline{1.3pt}
	        \multicolumn{2}{l}{CAD+}\\
	        \hline\hline
	        Label & CAD & \textsl{MA}\\
	        \hline
	        \#All & 105 & 134\\
	        \hline
	        Maj & .48 & .43\\
	        \hline
	        \texttt{RB} & .58  & .59 \\
	        \hline
	        \texttt{HR} & .77 & --\\
	        Ours & \textbf{.83} & \textbf{.89}\\
            \bhline{1.3pt}
        \end{tabular}
    $}
    \caption{Accuracy on the FraCaS, MED, SICK, HANS, and CAD+ datasets}
    \label{tab:accuracy}
\end{table}

\subsection{Results and discussion}
Table \ref{tab:accuracy} shows the experimental results on all datasets.
\textit{Maj} is the accuracy of the majority baseline and \textit{Ours} is the accuracy of our system.
For FraCaS, \textit{+rule} shows the accuracy achieved by the addition of hand-coded rules,
which correct the errors in POS tagging and lemmatization,
as described in \newcite{haruta-etal-2020-logical}.
Given that MED and HANS use binary labels (\LabelYes\ and \LabelUnk), for these two datasets we modify the system so that it outputs \LabelYes\ if the hypothesis can be proved from the premise; otherwise, the output is \LabelUnk.
We compare our system with previous logic-based systems and DL-based systems.

\paragraph{FraCaS}
We use three logic-based systems: \texttt{MN}~\cite{mineshima2015higher}, \texttt{LP}~\cite{abzianidze2016natural}, and \texttt{HR}~\cite{haruta-etal-2020-logical}.
These are systems based on CCG parsing and theorem proving.
For a DL-based system,
we use a state-of-the-art model,
RoBERTa (\texttt{RB})~\cite{Liu2019RoBERTaAR}, trained on MultiNLI~\cite{N18-1101},
using the implementation provided in AllenNLP~\cite{gardner-etal-2018-allennlp}.
Our system achieved nearly 100\% accuracy and outperformed the DL-system by a large margin.
FraCaS-241 in Table \ref{tab:data}
is a complex inference with numerical expressions;
this problem is solved by our system but neither by the other logic-based systems nor by the DL-based system (\texttt{RB}).
Our system also improved its predecessor (\texttt{HR})
in that it can handle inferences involving clausal comparatives (FraCaS 239--241).

\paragraph{MED}
For the results on MED in Table \ref{tab:accuracy},
\texttt{BERT} shows the performance of a BERT model fine-tuned with MultiNLI and \texttt{BERT+} shows that of a BERT model with data augmentation for monotonicity inferences in addition to the MultiNLI training set.
Both models were tested in \newcite{yanaka-etal-2019-neural}.
Our system outperformed both the logic-based system (\texttt{HR}) and the DL-based systems.
MED-176 and MED-485 in Table \ref{tab:data}, which
involve a downward quantifier (\textit{few}) and a non-monotonic quantifier (\textit{exactly 12}), respectively, are examples that our system correctly solved while the DL-models did not.

\paragraph{SICK}
For SICK, \texttt{MG}~\cite{martinez-gomez-etal-2017-demand} is a system based on CCG parsing with compositional event semantics and theorem proving.
Our system outperformed the DL-based system (\texttt{RB}) and
achieved comparable results with the logic-based systems (\texttt{LP} and \texttt{MG}), showing that 
the combination of event semantics and degree semantics is compatible with the insertion of lexical knowledge.
For example, SICK-1357 in Table \ref{tab:data} is an example involving the lexical inference from \textit{puppy} to \textit{dog};
our system correctly predicted the \LabelYes\ label for this problem, while the DL-based system (\texttt{RB}) predicted the \LabelNo\ label.

\paragraph{HANS}
\newcite{mccoy-etal-2019-right} reported that 
DL-based systems tend to erroneously output \LabelYes\ 
for cases in which the hypothesis was a constituent or a sub-string of the premise, such as disjunctive sentences (e.g., HANS-23991 in Table \ref{tab:data}).
To see how a system performs on these cases,
we present the accuracy for each gold answer label (\LabelYes\ and \LabelUnk).
While the accuracy when the gold label was \LabelYes~was close to 100\% in both our system and the DL-based system (\texttt{RB}),
the accuracy of our system was higher than that of \texttt{RB} when the label is \LabelUnk\ (78\% vs. 56\%).
One of the reasons for the relatively low accuracy (78\%) of our system in comparison with the performance on the other datasets is parse error; HANS contains syntactically complex sentences such as
\textit{The actor paid in the library recognized the lawyers} (HANS-13628, \textit{subsequence}),
for which the current CCG parsers output incorrect parses. Another reasons is inference involving a modal adverb, e.g.,
the inference from \textit{Probably the secretary admired the athlete} to \textit{The secretary admired the athlete} (HANS-24034). The gold label is \LabelUnk, but our system predicts \LabelYes, because any adverb can be dropped in the current implementation. A more fine-grained classification of adverbs will be needed to handle this type of inference.

\paragraph{CAD+}
For CAD+, our system outperformed the previous logic-based system (\texttt{HR}) and the DL-based system (\texttt{RB}).
Our system was able to solve the inference involving numerical computation (CAD-115) and antonym conversion for adverbs (CAD-157)
in Table \ref{tab:data}, while \texttt{RB} incorrectly predicted 
\LabelNo\ for CAD-115 and \LabelYes\ for CAD-157.
However, some problems with adverbial expressions remain.
For example, the sentence
\textit{Jones drives more carefully today than yesterday}
(\textsl{MA}-183) conjoins two adverbs \textit{today} and \textit{yesterday} by \textit{than}.
The current system does not derive
the correct logical form for this type of complex coordinate structure formed by \textit{than}-clauses.
A further improvement of CCG parsing would be needed to handle such complex coordinate constructions.

\section{Conclusion}
\label{sec:conlusion}

We have presented a logic-based NLI system that combines event semantics and degree semantics
and evaluated the system on various datasets containing semantically challenging inferences.
The results showed that the combination of event semantics and degree semantics
is viable and works well on the type of complex logical inferences for which standard DL-based systems
show poor performance.
This study contributes to the study of computational modeling and the evaluation of formal semantic theories,
as well as to the creation of challenging NLI problems that DL-based models need to address.

\paragraph{Acknowledgments}
We are grateful to the three anonymous reviewers.
This work was supported by JSPS KAKENHI Grant Number JP18H03284.

\bibliographystyle{coling}
\bibliography{coling2020}

\noautomath
\medskip
\end{document}